\documentclass{article}
\usepackage{spconf,amsmath,amssymb,graphicx,xcolor,comment}
\usepackage[pagebackref,breaklinks,colorlinks]{hyperref}

\title{ENACT: Entropy-based Clustering of Attention Input for Reducing\\ the Computational Needs of Object Detection Transformers}
\name{Giorgos Savathrakis$\rm{^1}$, Antonis Argyros$\rm{^{1,2}}$}
\address{$\rm{^1}$Institute of Computer Science, FORTH \hspace{0.5cm} $\rm{^{2}}$Computer Science Department, University of Crete\\
{\tt\small $\{$gsav, argyros$\}$@ics.forth.gr}}
\begin{document}
%\ninept
%
\maketitle
\begin{abstract}
Transformers demonstrate competitive performance in terms of precision on the problem of vision-based object detection. However, they require considerable computational resources due to the quadratic size of the attention weights. 
In this work, we propose to cluster the transformer input on the basis of its entropy, due to its similarity between same object pixels. This is expected to reduce GPU usage during training, while maintaining reasonable accuracy.
%%%AAA: Make sure that what is mentioed above is consistent with the final obtained results... 
This idea is realized with an implemented module that is called ENtropy-based Attention Clustering for detection Transformers (ENACT), which serves as a plug-in to any multi-head self-attention based transformer network. Experiments  on the COCO object detection dataset and three detection transformers demonstrate that the requirements on memory are reduced, while the detection accuracy is degraded only slightly. The code of the ENACT module is available at \url{https://github.com/GSavathrakis/ENACT}.
%%%AAA: Make sure that what is mentioed above is consistent with the final obtained results... 
\end{abstract}
\section{Introduction}
\label{sec:intro}
%%%AAA: I understand that you have to find space, so you need to cut text. However, the start of the paper and the introduction became quite abrupt...
Object detection is an important and challenging problem in computer vision, with a wide range of detectors, mostly Deep Learning based, having been proposed over the years. These consist of Convolutional Neural Network (CNN) backbones, and networks that aim to localize object regions and classes.

\subsection{Transformer-based object detectors}
The development of object detectors with a transformer-like architecture~\cite{10.1007/978-3-030-58452-8_13} resulted from the need to remove manually crafted components like NMS suppression or anchor generation, existing in previous detectors~\cite{7780460,NIPS2015_14bfa6bb}. Maintaining the CNN backbone from older architectures, the rest of the network follows an encoder-decoder architecture as proposed by Vaswani et al~\cite{vaswani2017attention}. Both the encoder and decoder layers include multi-head self-attention (MHSA) modules that quantify the correlation between a set of Queries and Keys, and match the resulting weights to a set of corresponding Values. All three originate from the CNN output in the encoder layers. The decoder input comes from learnable object queries and from the encoder output. Computing the attention has quadratic space and time complexity $\mathcal{O}$(N$^2$), which is a main limitation of this transformer architecture. Ideas aiming to solve this problem either restricted the number of pixels each pixel should focus on~\cite{zhu2021deformable}, or clustered pixels depending on the similarity of their features (e.g., ~\cite{DBLP:conf/bmvc/Zheng0ZL0021,vyas2020fast}).
\begin{figure*}[t]
  \centering
   \includegraphics[width=1.\linewidth]{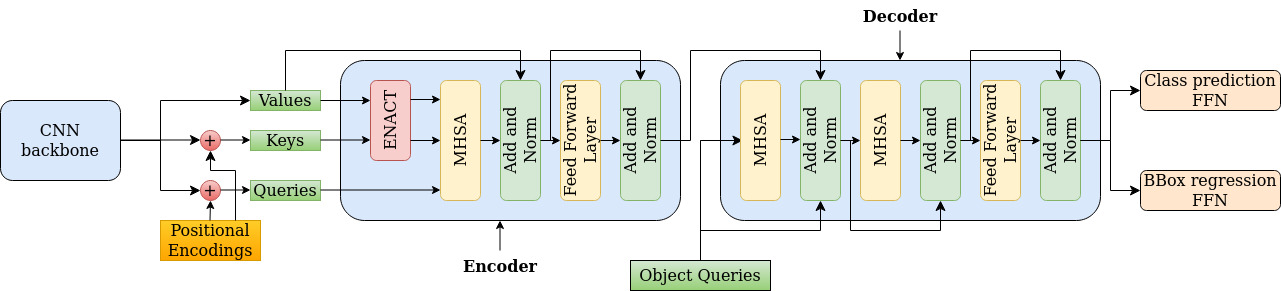}

   %\caption{Visualization of a detection transformer with the addition of the ENACT module. Queries, Keys and Values come from the CNN backbone, with the former two being supplemented with positional encodings, and the latter two being clustered by the ENACT module. All three then pass through the MHSA module.}
   \caption{Visualization of a detection transformer with the addition of the ENACT module. The feature maps obtained from the CNN backbone, are the Queries, Keys and Values, with the former two being supplemented with positional encodings. All three pass through the encoder, with the Keys and Values being clustered by the ENACT module. Subsequently, the Queries, as well as the clustered Keys and Values pass through the multi-head self-attention module (MHSA).}
   %, making it faster in computation and cheaper in memory usage.}
   \label{fig:1}
\end{figure*}
\subsection{Information-based clustering}
Most of the works that implemented attention clustering, used the feature vectors' Euclidean distance as a metric, assuming a connection between dot product similarity and object class. However, this approach is slow and requires domain-driven hyperparameter tuning (e.g., number of clusters).

To overcome these problems, we propose the Shannon entropy of the Keys' pixels as the clustering criterion, because pixels with similar information will very likely belong to the same object. Furthermore, the sum of the pixels' self-information is the information of the sum, which helps at merging pixels, while optimally retaining their information. 

Therefore, in this work, we propose the ENtropy-based Attention Clustering for detection Transformers (ENACT). ENACT clusters the Keys and Values by estimating the p.d.f of the Key input using a linear layer, computing the entropy using the p.d.f, and finally implementing grouping on the basis of the information signal's second derivative. The major contributions of our approach are the following:  

\begin{itemize}
    \item ENACT can serve as a {\sl plug-in to any detection transformer} architecture with an MHSA module in the encoder, and is orthogonal to any other memory reducing method (e.g., smaller arithmetic precision). Additionally, the clustering is based on the Key pixels' self-information similarity. Therefore, {\sl the number of clusters is not a hyper-parameter} and instead, its computation is data-driven.
    \item To the best of our knowledge, ENACT is the only method which handles variable-length tensors in the same batch, in a GPU parallelizable manner (see supplementary material for more details).
    \item We plug ENACT to three Detection Transformers~\cite{10.1007/978-3-030-58452-8_13,meng2021conditional,wang2022anchor} and we show that we obtain similar precision, while saving about 15-46$\%$ GPU RAM depending on the detector.
\end{itemize}
\section{RELATED WORK}
\label{sec:relatedwork}

Transformer networks were initially developed to deal with sequence translation problems (i.e. text-to-text translation etc.). Vaswani et al~\cite{vaswani2017attention} who introduced them used an encoder-decoder architecture where the input passes through a self-attention module in order to learn the connections between queries and keys. Dosovitskiy et al~\cite{dosovitskiy2021an} developed the Vision Transformer (ViT), considering Queries and Keys being image patches. Transformers for object detection were initially proposed by Carion et al~\cite{10.1007/978-3-030-58452-8_13} who developed the DETR model, which implements a bipartite matching loss that unilaterally connects a prediction to a ground truth object, in addition to a transformer-based encoder-decoder structure. This removed the need for Region Proposal Networks and NMS suppression, which are components required by older detectors such as YOLO~\cite{7780460} and Faster R-CNN~\cite{NIPS2015_14bfa6bb}. 

To solve the problem of the large space and time complexity Zhu et al proposed the Deformable DETR algorithm~\cite{zhu2021deformable}, where the attention of each pixel is focused on a predetermined number of points, with learnable attention weights and point locations. Other variants such as the Anchor DETR~\cite{wang2022anchor} and Conditional DETR~\cite{meng2021conditional} also propose methods that lead to reduced training times through modifications in the query design. 
%Another method that was proposed by Katharopoulos et al ~\cite{katharopoulos2020transformers} reduced the complexity of the attention calculation from quadratic to linear, using the associativity property of matrix products.

Improving transformers in terms of memory can also be done by clustering the input. Vyas et al ~\cite{vyas2020fast} used k-means to group queries into clusters, whereas works done by Zeng et al ~\cite{zeng2022not} and Haurum et al ~\cite{haurum2025agglomerative}, implemented hierarchical clustering, starting with many clusters and gradually reducing them to include the important regions. Zheng et al ~\cite{DBLP:conf/bmvc/Zheng0ZL0021} proposed the Adaptive Clustering Transformer (ACT) for DETR, which clusters the queries into prototypes using locality sensitive hashing to group queries with small Euclidean distance.

\section{METHOD}
\label{sec:Method}
\subsection{Overview}
\label{sec:3.1}

\begin{figure*}[t]
  \centering
   \includegraphics[width=0.85\linewidth]{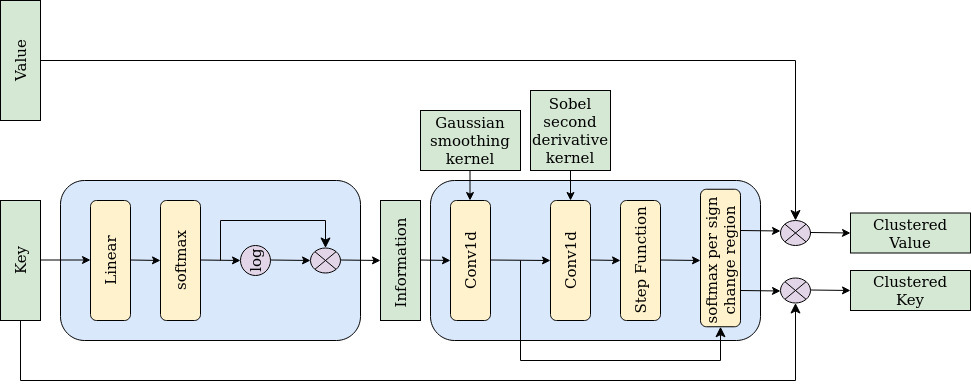}

   \caption{Overview of the ENACT module. The key passes through the network responsible for the calculation of the information and subsequently the information is smoothed by a Gaussian kernel. Then, the regions where the information is a convex or concave function are identified by the second derivative, and a softmax function is applied to the information values of each separate region of same sign in second derivative. The clustering is done by summing the weighted information values multiplied by the respective feature vectors of the values and keys, per region.}
   \label{fig:2}
\end{figure*}

In this work we propose the ENACT module, which is a plug-in clustering component to the standard detection transformer architecture (see Figure~\ref{fig:1}), based on each pixel's self-information. The information computation is data-driven and therefore, there is {\sl no pre-determination of the number of clusters}. Furthermore, it reduces the size of the input, which has a significant impact on the GPU memory required for training.

In detection transformers, the Queries (Q), Keys (K) and Values (V) are the same, with the exception that Queries and Keys are supplemented with positional encodings, that add information about the location of each pixel.
The dimensions of Q, K and V, are $\rm{N \times HW \times d}$ where $\rm{N}$ is the batch size, $\rm{HW}$ the total number of pixels in the feature map, and $\rm{d}$ the dimensions of the feature vector. Thus, computing the attention weights (see equation \ref{eq:1}), has quadratic complexity. By clustering the Keys and Values, the resulting $\rm{K_{cl}, V_{cl} \in \mathbb{R}^{H'W' \times d}}$ where $\rm{H'W' < NHW}$. This will reduce the size of the attention weights, and since the Values are also clustered, computing the final attention map (see equation \ref{eq:01}) is also more memory efficient.
%In the original detection transformers, Queries (Q), Keys (K) and Values (V), have the same dimensions, namely $\rm{Q, K, V \in \mathbb{R}^{N \times HW \times d}}$, where $N$ is the batch size, $HW$ the total number of pixels in the feature map, and $d$ the dimensions of the feature vector. 
%This means that the resulting attention weight matrix $\rm{A \in \mathbb{R}^{N \times HW \times HW}}$  is quadratic with respect to the spatial dimensions. In the ENACT model we cluster the Keys and the Values before passing them to the attention module, and the resulting $\rm{K_{cl}, V_{cl} \in \mathbb{R}^{N \times H'W' \times d}}$ where $\rm{H'W' < HW}$. Therefore, the clustered attention weight matrix $\rm{A_{cl} \in \mathbb{R}^{N \times HW \times H'W'}}$ is much less consuming in memory space. Additionally, since the Values are also clustered, the matrix multiplication between attention weights and values, which is done in order to compute the attention map ($\rm{\mathcal{A}}$), will be done along the axis of the reduced spatial dimensions. This will make the computation of the clustered attention map ($\rm{\mathcal{A}_{cl}}$) faster. To clarify this, in Eq.~\ref{eq:1} and Eq.~\ref{eq:01} we show how the attention weights and attention map, are calculated in the attention module.

\begin{equation}
  \rm{A = softmax \left(\frac{Q \cdot K^T}{\sqrt{d}}\right), \hspace{0.05cm} A_{cl} = softmax \left(\frac{Q \cdot K_{cl}^T}{\sqrt{d}}\right).}
  \label{eq:1}
\end{equation}

\begin{equation}
  \rm{\mathcal{A} = A \cdot V, \hspace{0.2cm} \mathcal{A}_{cl} = A_{cl} \cdot V_{cl}.}
  \label{eq:01}
\end{equation}

The Queries are left un-clustered because we want the final attention map to maintain its size after each encoder layer, in order to dimensionally match the positional encodings. 

The ENACT module (see Figure~\ref{fig:2}), consists of two main components. The first computes the self-information of the input pixels, and the second clusters regions of information gain and loss. To that end, we compute the entropy of the Keys because the positional encodings add locality information which influences the entropy. Also, it is plugged in the encoder only, since the decoder input is trainable and initialized as Gaussian noise. Therefore, clustering will damage training and the entropy of the noise will be meaningless.
%Since, we cluster the Keys and Values, we have to choose whose entropy will be the one on which the clustering will be based on. The input chosen for that are the Keys because they are supplemented with positional encodings, and we consider that the information that corresponds to the location of each feature vector is useful to the calculation of the entropy. This is because pixels with information regarding an object in an image, are likely to also be close to each other and the positional information is indicative of this proximity. Another important note is that we plug the ENACT model only inside the encoder of the transformer. This is because the input of the decoder is originally Gaussian noise and the computation of the pixels' self-information will not have any significant value. Another reason is that the decoder input is a trainable parameter, and clustering the input originating from Gaussian noise may be detrimental for the object query training.

\subsection{Clustering process}
The clustering is done as shown in Eq.~\ref{eq:2}. $\rm{\mathcal{H}(x)}$ is the self-information (i.e. the quantity inside the Shannon entropy sum), $\rm{x}$ is the feature vector of a pixel, and $\rm{p(x)}$ is the learnable p.d.f. of the vector distribution. The p.d.f. is learned using a linear layer $\rm{\mathbb{R}^{d\rightarrow 1}}$ and a softmax function.
%We can see that the computed quantity is the one inside the sum of the Shannon entropy.

\begin{equation}
  \rm{\mathcal{H}(x) = -p(x)log(p(x)).}
  \label{eq:2}
\end{equation}
%The p.d.f. is learnable since it is calculated by passing the input through a linear layer that maps it from the feature dimensions to 1 and the result is passed through a softmax function, making the sum of the probabilities across all pixels to be equal to one. This is shown in Eq.~\ref{eq:3}, where $W$ is the weight matrix, and $b$ is the bias. 

%\begin{equation}
  %\rm{p(x) = softmax(x\cdot W^T + b).}
  %\label{eq:3}
%\end{equation}
Therefore, the final output  of the information module is a signal with shape (batch size $\times$ $HW$) where $HW$ is the total number of pixels in the feature map.

Subsequently, we cluster K and V using the computed entropy, and a 1D convolution with a Gaussian kernel to smoothen the signal. The standard deviation $\rm{\sigma}$ is a hyperparameter and the kernel ranges from -3$\rm{\sigma}$ to 3$\rm{\sigma}$.
%Having obtained the information from the first module, we use it to cluster the attention Keys and Values. To do that we firstly convolve the information with a 1D Gaussian kernel. This is done because in the resulting information there may be cases with consecutive sharp edges in the signal due to noise, and in order to alleviate this problem, it is required to smooth the signal. The kernel itself is shown in Eq.~\ref{eq:4}, where the $x$ is a 1D vector of integers that range from -3$\rm{\sigma}$ to 3$\rm{\sigma}$:

After the smoothing is completed, we compute the regions where the signal is concave and convex. To do so, we first estimate its second derivative using the Sobel kernel in Eq.~\ref{eq:5}.
\begin{equation}
    \rm{Sobel''(x) = [-1, 2, -1].}
    \label{eq:5}
\end{equation}
The reason we do this is because we consider the regions where the local information gain or loss is apparent as the most suitable for clustering, since, for example, a pixel with an information local maximum/minimum and its surroundings could be clustered as one entity.

Subsequently, we pass the second derivative of the information to a step function which attributes 1 to the convex parts of the information curve, and -1 to the concave ones. Concave regions signify information gain, and convex ones signify loss.

At that point, we take the smoothed entropy and we run a softmax function on each separate region whose indices correspond to regions of the same sign in the output of the step function. 
%For example, assume that at one point the output of the step function is $[...1,1,1,-1,-1,1,1,1,1, ...]$ and the indices are $[...i, i+1,...,i+8,...]$. In the respective indices of the entropy, the softmax function will be run on three separate regions, which are from $i$ to $i+2$, $i+3$ to $i+4$ and $i+5$ to $i+8$. 
The final step is to multiply the resulting weighted entropies with the respective feature vectors from the Keys and Values, and sum the results per cluster. To achieve this goal, we implement methods that take advantage of GPU parallelization by creating CUDA kernels that make the necessary operations in an optimal manner. Further information is provided in the  supplementary material\footnote{ \url{https://sigport.org/sites/default/files/docs/ICIP_2025_Supplementary_Material.pdf}}.
\begin{table*}[t]
    \centering
    \begin{tabular}{@{}l|c|c|c|c@{}} \hline
    %\toprule
        
        %\midrule
        
        \textbf{Model} & \textbf{GPU RAM (GB)} & \textbf{GPU Gain ($\%$)} & \textbf{training time (s/image)} & \textbf{Inference time (s/image)} \\ \hline
        DETR~\cite{10.1007/978-3-030-58452-8_13} & 36.5 & -  & \textbf{0.0541} & \textbf{0.0482}  \\
        DETR + ACT~\cite{DBLP:conf/bmvc/Zheng0ZL0021} & 28.9 & 20.8 & 0.0634 & 0.0494 \\
        DAB-DETR~\cite{liu2022dabdetr} & 26.1 & 28.5 & 0.0965 & 0.0590 \\
        MS-DETR~\cite{zhao2024ms} & 21.0 & 42.5 & 0.1557 & 0.1088 \\
        DETR + ENACT & \textbf{19.7} & \textbf{46.0} & 0.0693 & 0.0538 \\ \hline 
        Anchor DETR~\cite{wang2022anchor} & 29.7 & - & \textbf{0.0999} & \textbf{0.0707} \\
        Anchor DETR + ENACT & \textbf{25.1} & 15.5 & 0.1170 & 0.0779 \\ \hline
        Conditional DETR~\cite{meng2021conditional} & 46.6 & - & \textbf{0.0826} & \textbf{0.0646} \\
        Conditional DETR + ENACT & \textbf{37.3} & 20.0 & 0.0930 & 0.0693 \\ \hline
        %\bottomrule
      \end{tabular}
      \caption{Comparison of GPU RAM consumption in GB, training and inference time in seconds per image, for three detection transformers, with and without the plugging of the ENACT module. We also show the GPU memory percentage gain when using ENACT. For DETR we also provide the comparisons with the ACT clustering module.}
      \label{tab:4.5.1}
\end{table*}
\begin{table*}[t]
  \centering
  \begin{tabular}{@{}l|c|c c c c c c@{}} \hline
    %\toprule
    
    %\midrule
    
    \textbf{Model} & \textbf{Epochs} & \textbf{AP} & \textbf{AP$\rm{_{50}}$} & \textbf{AP$\rm{_{75}}$} & \textbf{AP$\rm{_s}$} & \textbf{AP$\rm{_m}$} & \textbf{AP$\rm{_l}$}\\ \hline
    RetinaNet~\cite{lin2017focal} & 36 & 38.7 & 58.0 & 41.5 & 23.3 & 42.3 & 50.3 \\
    Faster RCNN~\cite{NIPS2015_14bfa6bb} & 36 & 40.2 & 61.0 & 43.8 & 24.2 & 43.5 & 52.0 \\
    RSDNet$\rm{^{\dag}}$ ~\cite{wang2020rdsnet} & 12 & 40.3 & 60.1 & 43.0 & 22.1 & 43.5 & 51.5 \\
    Cascade R-CNN$\rm{^{\dag}}$ ~\cite{cai2018cascade} & 12 & 42.7 & 61.6 & 46.6 & 23.8 & 46.2 & 57.4 \\
    \hline \hline
    DETR-C5~\cite{10.1007/978-3-030-58452-8_13} & 300 & 40.6 & 61.6 & - & 19.9 & 44.3 & 60.2 \\
    %DETR + ACT~\cite{DBLP:conf/bmvc/Zheng0ZL0021} & 300 & 40.5 & 60.9 & 42.6 & 19.1 & 44.5 & 59.8 \\
    %DAB-DETR~\cite{liu2022dabdetr} & 50 & 42.2 & 63.1 & 44.7 & 21.5 & 45.7 & 60.3 \\
    %MS-DETR~\cite{zhao2024ms} & 12 & 48.8 & 66.2 & 53.2 & 31.5 & 52.3 & 63.7 \\
    DETR-C5 + ENACT (ours) & 300 & 40.1 & 60.5 & 41.9 & 19.4 & 43.5 & 58.2 \\ 
\hline
    Anchor DETR-DC5~\cite{wang2022anchor} & 50 & 44.3 & 64.9 & 47.7 & 25.1 & 48.1 & 61.1 \\
    Anchor DETR-DC5 + ENACT (ours) & 50 & 42.9 & 63.4 & 46.2 & 24.6 & 46.9 & 58.5 \\ \hline

    Conditional DETR-C5$\rm{^{\dag}}$ ~\cite{meng2021conditional} & 50 & 42.8 & 63.7 & 46.0 & 21.7 & 46.6 & 60.9 \\ 
    Conditional DETR-C5 + ENACT$\rm{^{\dag}}$ (ours) & 50 & 41.5 & 62.6 & 44.1 & 21.9 & 45.3 & 59.2 \\ \hline
    %\bottomrule
  \end{tabular}
  \caption{Detection performance results in terms of precision, as well as number of training epochs, for different object detectors, on the COCO val2017. DC5 and C5 denote whether the last convolutional layer in the DETR based models is dilated or not, respectively. All models share the ResNet-50 backbone, unless denoted by $\rm{^{\dag}}$ which means they were trained using ResNet-101 backbone. (AP$\rm{_{75}}$ not available in DETR 300 epoch schedule).}
  \label{tab:4.5.3}
\end{table*}
%------------------------------------------------------------------------
\section{EXPERIMENTS}
\label{sec:Experiments}
\subsection{Dataset and Settings}
\label{sec:4.2}
We evaluate our model on the MS COCO 2017 dataset, using the train2017 subset for training, and the val2017 for validation.  
%~\cite{lin2014microsoft}, which is a benchmark dataset for object detection. The 2017 version of the dataset is split into train, validation and test sets, which consist of 118,000, 5,000 and 41,000 images, respectively. Since annotations are not available for the test set, we use only the training part of the dataset for training, and we evaluate the overall performance on the validation set.

As it was mentioned in Section~\ref{sec:3.1}, the Keys and Values pass through the ENACT module for dimensional consistency reasons, and the entropy is computed only from the Keys. The parameters are initialized from a Xavier uniform distribution and the biases are initially zero.
\begin{figure*}[htb]

\begin{minipage}[b]{0.19\linewidth}
  \centering
  \centerline{\includegraphics[width=1.\linewidth]{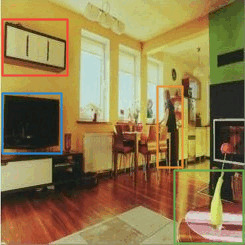}}
%  \vspace{2.0cm}
\end{minipage}
\begin{minipage}[b]{0.19\linewidth}
  \centering
  \centerline{\includegraphics[width=1.\linewidth]{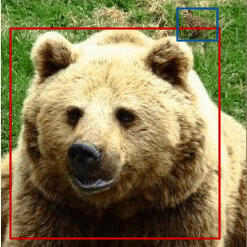}}
%  \vspace{2.0cm}
\end{minipage}
\begin{minipage}[b]{0.19\linewidth}
  \centering
  \centerline{\includegraphics[width=1.\linewidth]{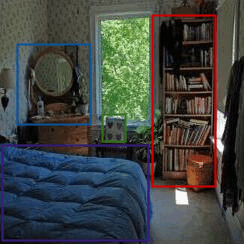}}
%  \vspace{2.0cm}
\end{minipage}
\begin{minipage}[b]{0.19\linewidth}
  \centering
  \centerline{\includegraphics[width=1.\linewidth]{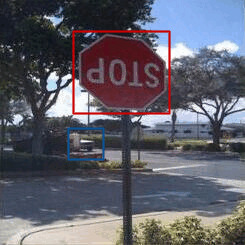}}
%  \vspace{2.0cm}
\end{minipage}
\begin{minipage}[b]{0.19\linewidth}
  \centering
  \centerline{\includegraphics[width=1.\linewidth]{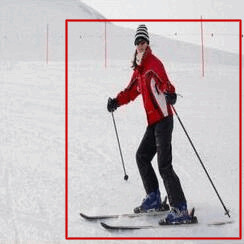}}
%  \vspace{2.0cm}
\end{minipage}
\hfill
\begin{minipage}[b]{0.19\linewidth}
  \centering
  \centerline{\includegraphics[width=1.\linewidth]{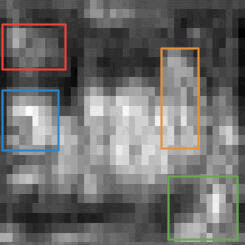}}
%  \vspace{2.0cm}
\end{minipage}
\hspace{0.07cm}
\begin{minipage}[b]{0.19\linewidth}
  \centering
  \centerline{\includegraphics[width=1.\linewidth]{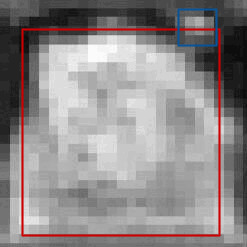}}
%  \vspace{2.0cm}
\end{minipage}
\hspace{0.07cm}
\begin{minipage}[b]{0.19\linewidth}
  \centering
  \centerline{\includegraphics[width=1.\linewidth]{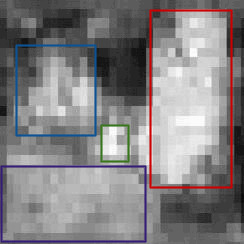}}
%  \vspace{2.0cm}
\end{minipage}
\hspace{0.07cm}
\begin{minipage}[b]{0.19\linewidth}
  \centering
  \centerline{\includegraphics[width=1.\linewidth]{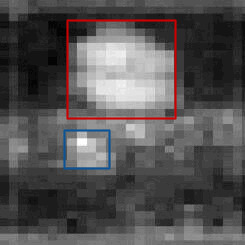}}
%  \vspace{2.0cm}
\end{minipage}
\hspace{0.07cm}
\begin{minipage}[b]{0.19\linewidth}
  \centering
  \centerline{\includegraphics[width=1.\linewidth]{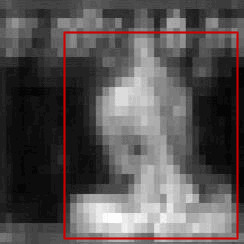}}
%  \vspace{2.0cm}
\end{minipage}
\caption{A selection of images from the COCO dataset (above), with the respective output of the self-information module prior to the Gaussian smoothing (below). Bright pixels in the bottom images, correspond to regions that yield higher information. We use same-color bounding boxes, to show corresponding objects between images and information maps.}
\label{fig:4}
\end{figure*}
We plug ENACT to three Detection Transformers, which are the DETR~\cite{10.1007/978-3-030-58452-8_13}, Conditional DETR~\cite{meng2021conditional} and Anchor DETR~\cite{wang2022anchor}, and we use the ResNet 50 and 101 backbones~\cite{7780459}.
%, the AdamW optimizer~\cite{loshchilov2018decoupled} with $5e-5$ learning rate and a drop schedule which reduces its value to a tenth of the original. 
DETR is trained for 300 epochs, dropping the learning rate at 200, and the other two for 50 epochs with the drop at 40. The batch size is set to 4 for the Anchor-DETR and 8 for the other two transformers. For our experiments, we used an NVIDIA RTX A6000 GPU with 49GB RAM. 

The standard deviation $\rm{\sigma}$ of the Gaussian smoothing kernel, is set to 5 for DETR and Conditional DETR, and to 3 for Anchor DETR. %In the case of Anchor DETR, which uses Row-Column Decoupled Attention (RCDA), we compute the entropy based on the Key rows, since the regions of information gain and loss are almost identical whether we use the Key rows or columns (see Section~\ref{sec:abl_st}). 
To evaluate the proposed ENACT module, we report GPU memory usage, training and inference time, and average precision (AP), as well as average precision at different IoU thresholds and different object area sizes.

\subsection{Ablation studies}
\label{sec:abl_st}

The only tunable hyperparameter is the standard deviation of the Gaussian smoothing kernel. It is determined by the AP obtained after a certain number of epochs for each detector. For DETR, after 10 epochs of training, we get an AP of 12.8, 12.9 and 12.7 for $\rm{\sigma=3,5,7}$ respectively. For Anchor-DETR we get 25.5 and 25.1 AP at epoch 5 for $\rm{\sigma=3,5}$ respectively, and finally for Conditional-DETR at epoch 5 we get AP 18.6 and 19.4 for $\rm{\sigma=3,5}$ respectively. Therefore, $\rm{\sigma}$ is set to 5, 3 and 5 for the respective detectors.
%With respect to hyperparameter tuning in the ENACT module, we only have to determine the standard deviation of the Gaussian smoothing kernel as mentioned in Section~\ref{sec:4.2}. The optimal decision varies per detector used, and depends on the value that leads to maximum precision and maximum gain in memory and training time. We observe that for $\rm{\sigma=3}$ we get the maximum AP at epoch 10, even higher than the one of the plain DETR at the same epoch. A similar approach is followed for determining the optimal $\rm{\sigma}$ for Anchor DETR and Conditional DETR. In the case of Anchor DETR, at epoch 2, $\rm{\sigma=3}$ yielded AP 15.3, whereas $\rm{\sigma=5}$ yielded AP 16.4 with a significant reduction on GPU memory of $\rm{\sim 50\%}$. Therefore, for Anchor DETR, we set $\rm{\sigma=5}$. As mentioned previously, Anchor DETR implements row-column decoupling and we only use the Key rows for the computation of the entropy, which as we said, leads to almost identical regions. Specifically, we observe that the cluster regions resulting from using the Key rows and then the Key columns for the computation of the entropy are similar by 95$\%$ on average. Finally, regarding Conditional DETR, at epoch 2, $\rm{\sigma=3}$ yielded AP 12.36, whereas $\rm{\sigma=5}$ yielded AP 12.34, and the GPU memory usage was similar. Therefore, we kept $\rm{\sigma=3}$ for the remainder of the training.

\subsection{Results}
\noindent{\bf Quantitative results:}
We present the overall performance of the ENACT module, when plugged into the three mentioned detection transformers. Starting with the GPU requirements, in Table~\ref{tab:4.5.1} we show the RAM consumption of the GPU when using the three detection transformer variants, and when plugging in the ENACT module to each of them. In the case of DETR, we also provide results for DAB-DETR~\cite{liu2022dabdetr}, MS-DETR~\cite{zhao2024ms} and the ACT module~\cite{DBLP:conf/bmvc/Zheng0ZL0021}. They are all algorithms that tweak the transformer input, and we present them in order to compare the respective gains in memory and time.
%In the case of DETR, we also provide results for the ACT module~\cite{DBLP:conf/bmvc/Zheng0ZL0021} which is another clustering algorithm, in order to compare the respective gains in memory and time. 
%The results are those obtained during training so that both forward pass inputs and backward pass gradients are included. 
We can observe that by supplementing each of the detection transformers with the ENACT module, we consistently reduce the GPU consumption by a margin which ranges from 15$\%$ to 46$\%$. In addition, compared to ACT, DAB-DETR and MS-DETR, we surpass them with respect to GPU gain by 26$\%$, 17.5$\%$ and 3.5$\%$ respectively. We note that the GPU memory values for DAB-DETR and MS-DETR are conservative estimates, obtained by adjusting the values claimed by the respective authors, to our batch size. Actual experiments using these two detectors, may yield GPU memory values that are larger than the ones presented in this paper.
%This constitutes a significant reduction in computational resources, which can extend the deployment of such transformer models by smaller GPUs. 

%Subsequently, we evaluate the performance of the ENACT module with regard to its capacity of reducing the total training time required by the model upon which it is applied. In the same manner as for the GPU memory usage, we use the models with and without ENACT, and we compare the time per image required in both cases. The results are also shown in Table \ref{tab:4.5.1}, where we show the time per image required in both cases for the same three detection transformers. It can be verified that the training time required is reduced by 5.6$\%$ in the Conditional DETR model, 9.8$\%$ in the DETR model and 15.4$\%$ in the Anchor DETR model. Therefore,  the reduction of the Keys and Values sizes used in the self-attention function, indeed reduces the time required for the dot product operations. Moreover, it is observed that the gain obtained in terms of inference time ranges from 2$\%$ in the DETR, to 14$\%$ in the Anchor DETR. 

Finally, we evaluate ENACT's performance
%the performance obtained by the plugging of the ENACT module 
in terms of precision and present the results in Table~\ref{tab:4.5.3}. We observe that the obtained APs are very close to those of the original transformer models, with the reductions being 0.5$\%$ for the DETR, 1.3$\%$ for the Conditional DETR and 1.4$\%$ for the Anchor DETR. Those are fairly slight drops considering the GPU gains, and the resulting precisions still surpass several existing object detectors in the same dataset.

The only cost comes in terms of training and inference time, where the delay ranges between 11 $\%$ to 22$\%$ in training time, and between 6 $\%$ to 10$\%$ in inference time.
%------------------------------------------------------------------------

\vspace*{0.2cm}\noindent{\bf Qualitative results:}
%One of the important aspects of this work that has to be verified, is the 
Subsequently, we verify the extent to which the idea of computing the self-information of each pixel is valid. Besides verifying that ENACT drops the needed computational resources without affecting considerably the detection precision, it is also important to visualize the output from ENACT's self-information component. Some indicative results are shown in Figure~\ref{fig:4}. Specifically, we plug the ENACT module in the Conditional DETR, and we pass the images through the trained model of the final epoch. We observe that for the most part, the visualized information map is brighter in regions where objects are located, and dimmer in the background. Considering the fact that the probability density function is learnable, we can see that the final trained model, in the part responsible for computing the information, correctly learned that the most informative pixels are the ones that correspond to objects of interest.

\section{SUMMARY}
\label{sec:Conclusions}
We  presented ENACT, a clustering module for MHSA-based detection transformers, which compresses the encoder input on the basis of its entropy. We showed that the pixel-wise self-information is indeed correlated to the important objects in the image, thereby rendering it as a valid metric for clustering. We trained three detection transformers using ENACT, and we show that we can obtain approximately 15$\%$ to 45$\%$ decrease in GPU memory, while losing only 0.5$\%$ to 1.4$\%$ in average precision. This makes ENACT a reliable plug-in for this type of transformer models, which can reduce significantly their computational requirements without compromising considerably their object detection capabilities.

\section{Acknowledgments}
% This work was co-funded by the Hellenic Foundation for Research and Innovation (HFRI) under the ``1st Call for HFRI Research Projects to support Faculty members and Researchers and the procurement of high-cost research equipment", project I.C.Humans, no 91.
This work was co-funded by the European Union (EU - HE Magician – Grant Agreement 101120731). The authors also gratefully acknowledge the support of the framework of the National Recovery and Resilience Plan Greece 2.0, funded by the European Union– NextGenerationEU (project Greece4.0).

%\vfill\pagebreak

\bibliographystyle{IEEEbib}
\bibliography{strings,refs,egbib}

\begin{thebibliography}{10}

\bibitem{10.1007/978-3-030-58452-8_13}
N.~Carion, F.~Massa, G.~Synnaeve, N.~Usunier, A.~Kirillov, and S.~Zagoruyko,
\newblock ``End-to-{E}nd {O}bject {D}etection with {T}ransformers,''
\newblock in {\em ECCV}. 2020, pp. 213--229, Springer International Publishing.

\bibitem{7780460}
J.~Redmon, S.~Divvala, R.~Girshick, and A.~Farhadi,
\newblock ``You {O}nly {L}ook {O}nce: {U}nified, {R}eal-{T}ime {O}bject {D}etection,''
\newblock in {\em 2016 IEEE Conference on Computer Vision and Pattern Recognition (CVPR)}, Los Alamitos, CA, USA, jun 2016, pp. 779--788, IEEE Computer Society.

\bibitem{NIPS2015_14bfa6bb}
Shaoqing Ren, Kaiming He, Ross Girshick, and Jian Sun,
\newblock ``Faster {R-CNN}: {T}owards {R}eal-{T}ime {O}bject {D}etection with {R}egion {P}roposal {N}etworks,''
\newblock in {\em Advances in Neural Information Processing Systems}, C.~Cortes, N.~Lawrence, D.~Lee, M.~Sugiyama, and R.~Garnett, Eds. 2015, vol.~28, Curran Associates, Inc.

\bibitem{vaswani2017attention}
Ashish Vaswani, Noam Shazeer, Niki Parmar, Jakob Uszkoreit, Llion Jones, Aidan~N Gomez, {\L}ukasz Kaiser, and Illia Polosukhin,
\newblock ``Attention {I}s {A}ll {Y}ou {N}eed,''
\newblock {\em Advances in Neural Information Processing Systems}, vol. 30, 2017.

\bibitem{zhu2021deformable}
Xizhou Zhu, Weijie Su, Lewei Lu, Bin Li, Xiaogang Wang, and Jifeng Dai,
\newblock ``Deformable {DETR}: {D}eformable {T}ransformers for {E}nd-to-{E}nd {O}bject {D}etection,''
\newblock in {\em International Conference on Learning Representations}, 2021.

\bibitem{DBLP:conf/bmvc/Zheng0ZL0021}
Minghang Zheng, Peng Gao, Renrui Zhang, Kunchang Li, Hongsheng Li, and Hao Dong,
\newblock ``End-to-{E}nd {O}bject {D}etection with {A}daptive {C}lustering {T}ransformer,''
\newblock in {\em 32nd British Machine Vision Conference 2021, {BMVC} 2021, Online, November 22-25, 2021}. 2021, p. 226, {BMVA} Press.

\bibitem{vyas2020fast}
Apoorv Vyas, Angelos Katharopoulos, and Fran{\c{c}}ois Fleuret,
\newblock ``Fast {T}ransformers with {C}lustered {A}ttention,''
\newblock {\em Advances in Neural Information Processing Systems}, vol. 33, pp. 21665--21674, 2020.

\bibitem{meng2021conditional}
Depu Meng, Xiaokang Chen, Zejia Fan, Gang Zeng, Houqiang Li, Yuhui Yuan, Lei Sun, and Jingdong Wang,
\newblock ``Conditional {DETR} for {F}ast {T}raining {C}onvergence,''
\newblock in {\em Proceedings of the IEEE/CVF International Conference on Computer Vision}, 2021, pp. 3651--3660.

\bibitem{wang2022anchor}
Yingming Wang, Xiangyu Zhang, Tong Yang, and Jian Sun,
\newblock ``Anchor {DETR}: {Q}uery {D}esign for {T}ransformer-{B}ased {D}etector,''
\newblock in {\em Proceedings of the AAAI Conference on Artificial Intelligence}, 2022, vol.~36, pp. 2567--2575.

\bibitem{dosovitskiy2021an}
Alexey Dosovitskiy, Lucas Beyer, Alexander Kolesnikov, Dirk Weissenborn, Xiaohua Zhai, Thomas Unterthiner, Mostafa Dehghani, Matthias Minderer, Georg Heigold, Sylvain Gelly, Jakob Uszkoreit, and Neil Houlsby,
\newblock ``An {I}mage is {W}orth 16x16 {W}ords: {T}ransformers for {I}mage {R}ecognition at {S}cale,''
\newblock in {\em International Conference on Learning Representations}, 2021.

\bibitem{zeng2022not}
Wang Zeng, Sheng Jin, Wentao Liu, Chen Qian, Ping Luo, Wanli Ouyang, and Xiaogang Wang,
\newblock ``Not {A}ll {T}okens are {E}qual: {H}uman-{C}entric {V}isual {A}nalysis via {T}oken {C}lustering {T}ransformer,''
\newblock in {\em Proceedings of the IEEE/CVF Conference on Computer Vision and Pattern Recognition}, 2022, pp. 11101--11111.

\bibitem{haurum2025agglomerative}
Joakim~Bruslund Haurum, Sergio Escalera, Graham~W Taylor, and Thomas~B Moeslund,
\newblock ``Agglomerative {T}oken {C}lustering,''
\newblock in {\em European Conference on Computer Vision}. Springer, 2025, pp. 200--218.

\bibitem{liu2022dabdetr}
Shilong Liu, Feng Li, Hao Zhang, Xiao Yang, Xianbiao Qi, Hang Su, Jun Zhu, and Lei Zhang,
\newblock ``{DAB}-{DETR}: {D}ynamic {A}nchor {B}oxes are {B}etter {Q}ueries for {DETR},''
\newblock in {\em International Conference on Learning Representations}, 2022.

\bibitem{zhao2024ms}
Chuyang Zhao, Yifan Sun, Wenhao Wang, Qiang Chen, Errui Ding, Yi~Yang, and Jingdong Wang,
\newblock ``{MS-DETR}: {E}fficient {DETR} {T}raining with {M}ixed {S}upervision,''
\newblock in {\em Proceedings of the IEEE/CVF Conference on Computer Vision and Pattern Recognition}, 2024, pp. 17027--17036.

\bibitem{lin2017focal}
Tsung-Yi Lin, Priya Goyal, Ross Girshick, Kaiming He, and Piotr Doll{\'a}r,
\newblock ``Focal {L}oss for {D}ense {O}bject {D}etection,''
\newblock in {\em Proceedings of the IEEE International Conference on Computer Vision}, 2017.

\bibitem{wang2020rdsnet}
Shaoru Wang, Yongchao Gong, Junliang Xing, Lichao Huang, Chang Huang, and Weiming Hu,
\newblock ``{RDSN}et: {A} {N}ew {D}eep {A}rchitecture for {R}eciprocal {O}bject {D}etection and {I}nstance {S}egmentation,''
\newblock in {\em Proceedings of the AAAI Conference on Artificial Intelligence}, 2020, vol.~34, pp. 12208--12215.

\bibitem{cai2018cascade}
Zhaowei Cai and Nuno Vasconcelos,
\newblock ``Cascade {R-CNN}: {D}elving into {H}igh {Q}uality {O}bject {D}etection,''
\newblock in {\em Proceedings of the IEEE Conference on Computer Vision and Pattern Recognition}, 2018, pp. 6154--6162.

\bibitem{7780459}
Kaiming He, Xiangyu Zhang, Shaoqing Ren, and Jian Sun,
\newblock ``Deep {R}esidual {L}earning for {I}mage {R}ecognition,''
\newblock in {\em 2016 IEEE Conference on Computer Vision and Pattern Recognition (CVPR)}, 2016, pp. 770--778.

\end{thebibliography}

\end{document}